\documentclass{article}

\usepackage[preprint]{neurips_2026}

\usepackage[utf8]{inputenc} % allow utf-8 input
\usepackage[T1]{fontenc}    % use 8-bit T1 fonts
\usepackage{hyperref}       % hyperlinks
\usepackage{url}            % simple URL typesetting
\usepackage{booktabs}       % professional-quality tables
\usepackage{amsfonts}       % blackboard math symbols
\usepackage{nicefrac}       % compact symbols for 1/2, etc.
\usepackage{microtype}      % microtypography
\usepackage{xcolor}         % colors
\usepackage{graphicx}

\usepackage{booktabs} % For better table lines
\usepackage{multirow} % For merging cells
\usepackage{graphicx}
\usepackage[usestackEOL]{stackengine} % 用于 \Centerstack 换行

\usepackage[table]{xcolor}

\usepackage{algorithm}
\usepackage{algpseudocode}
\usepackage{amsmath}
\usepackage{makecell} % 用于表头换行
\usepackage{arydshln} % 用于绘制虚线

\usepackage{makecell}
\usepackage{booktabs}
\usepackage[most]{tcolorbox}
\usepackage{etoc} % Local table of contents (useful for Supplementary Materials)
\definecolor{myLightBlue}{HTML}{E6F2FF} % 定义你之前的浅蓝色
\usepackage{tabularx}
\usepackage{enumitem}
\usepackage{collcell}             % 收集单元格内容用于计算
\usepackage{pgf}                  % 数学计算
\usepackage{etoc} % Local table of contents (useful for Supplementary Materials)
\title{\textit{EvalVerse}: Pipeline-Aware and Expert-Calibrated Benchmarking for Professional Cinematic Video Generation}

\author{%
  Songlin Yang$^{1,2,\dagger,}$\thanks{Project leader}~~, 
  Haobin Zhong$^{2,\dagger}$, 
  Ruilin Zhang$^{3,}$\thanks{Equal contribution}~~,\\ 
  \textbf{Xiaotong Zhao}$^2$\textbf{,} 
  \textbf{Shuai Li}$^2$\textbf{,} \textbf{Kai Zheng}$^2$\textbf{,} \textbf{Xuyi Yang}$^{1}$\textbf{,} \textbf{Zhe Wang}$^{1,2}$\textbf{,} \\\textbf{Zhenchen Tang}$^{2,4}$\textbf{,} \textbf{Yang Li}$^{2,4}$\textbf{,}
  \textbf{Bohai Gu}$^{1,2}$\textbf{,} \textbf{Zhengwei Peng}$^2$\textbf{,} \textbf{Yidan Huang}$^5$\textbf{,} \\\textbf{Mengzhou Luo}$^5$\textbf{,} \textbf{Yihang Bo}$^5$\textbf{,} \textbf{Dalu Feng}$^5$\textbf{,}
  \textbf{Yujia Zhang}$^2$\textbf{,} \textbf{Juntao Ma}$^2$\textbf{,} \textbf{Ruiqi Wang}$^2$\textbf{,}  \\ \textbf{Lvmin Zhang}$^6$\textbf{,} \textbf{Yuwei Guo}$^7$\textbf{,} \textbf{Frank Guan}$^8$\textbf{,} \textbf{Maneesh Agrawala}$^6$\textbf{,} \textbf{Hongbo Fu}$^1$\textbf{,} \\ 
  \textbf{Alan Zhao}$^2$\textbf{,} \textbf{Anyi Rao}$^{1,}$\thanks{Corresponding author} \\
  $^1$The Hong Kong University of Science and Technology, $^2$Tencent, $^3$Tsinghua University,\\ 
  $^4$Institute of Automation, Chinese Academy of Sciences, $^5$Beijing Film Academy, \\
  $^6$Stanford University, $^7$The Chinese University of Hong Kong, $^8$Singapore Institute of Technology \\
  \texttt{syangds@connect.ust.hk, anyirao@ust.hk}\\}

\begin{document}

\maketitle

\begin{abstract}
The rapid evolution of generative video foundation models has propelled the field toward professional-grade cinematic synthesis. To achieve such demanding quality, the community transitions towards Reinforcement Learning (RL) and agentic workflows. However, reliable evaluation has emerged as a critical bottleneck. Existing benchmarks predominantly evaluate ``whether it is right'' (basic prompt-following) while fundamentally neglecting ``whether it is good'' (cinematic quality, acting, and aesthetics). Furthermore, current automated metrics lack the domain-specific rigor required to provide trustworthy signals, creating a severe credibility gap between human aesthetic perception and machine scoring. To bridge this gap, we introduce \textbf{EvalVerse}, a comprehensive, pipeline-aware, and expert-calibrated evaluation framework. We treat video generation assessment not merely as an engineering task, but as a core scientific problem: the systematic digitization of subjective cinematic expertise. First, we \textbf{\textit{organize}} domain knowledge into an evaluation taxonomy aligned with the professional filmmaking workflow (pre-production, production, and post-production). Second, we \textbf{\textit{distill}} human expert judgments into a curated dataset with large-scale human annotations. Third, we \textbf{\textit{inject}} this knowledge into Vision-Language Models (VLMs) through an expert-calibrated fine-tuning strategy, enabling the VLM to perform explicit Chain-of-Thought reasoning. Compared to previous works, EvalVerse not only retains compatibility with foundational ``rightness'' metrics, but also significantly expands the criteria to ``goodness'' and broaden the task coverage to complex multi-shot sequencing and audio-visual integration. Consequently, by providing granular diagnostic signals, EvalVerse transcends a static leaderboard and establishes a fundamental infrastructure for future work, such as reward models and evaluator agent.
\end{abstract}

\begin{figure*}[ht]
\centering
  \includegraphics[width=\textwidth]{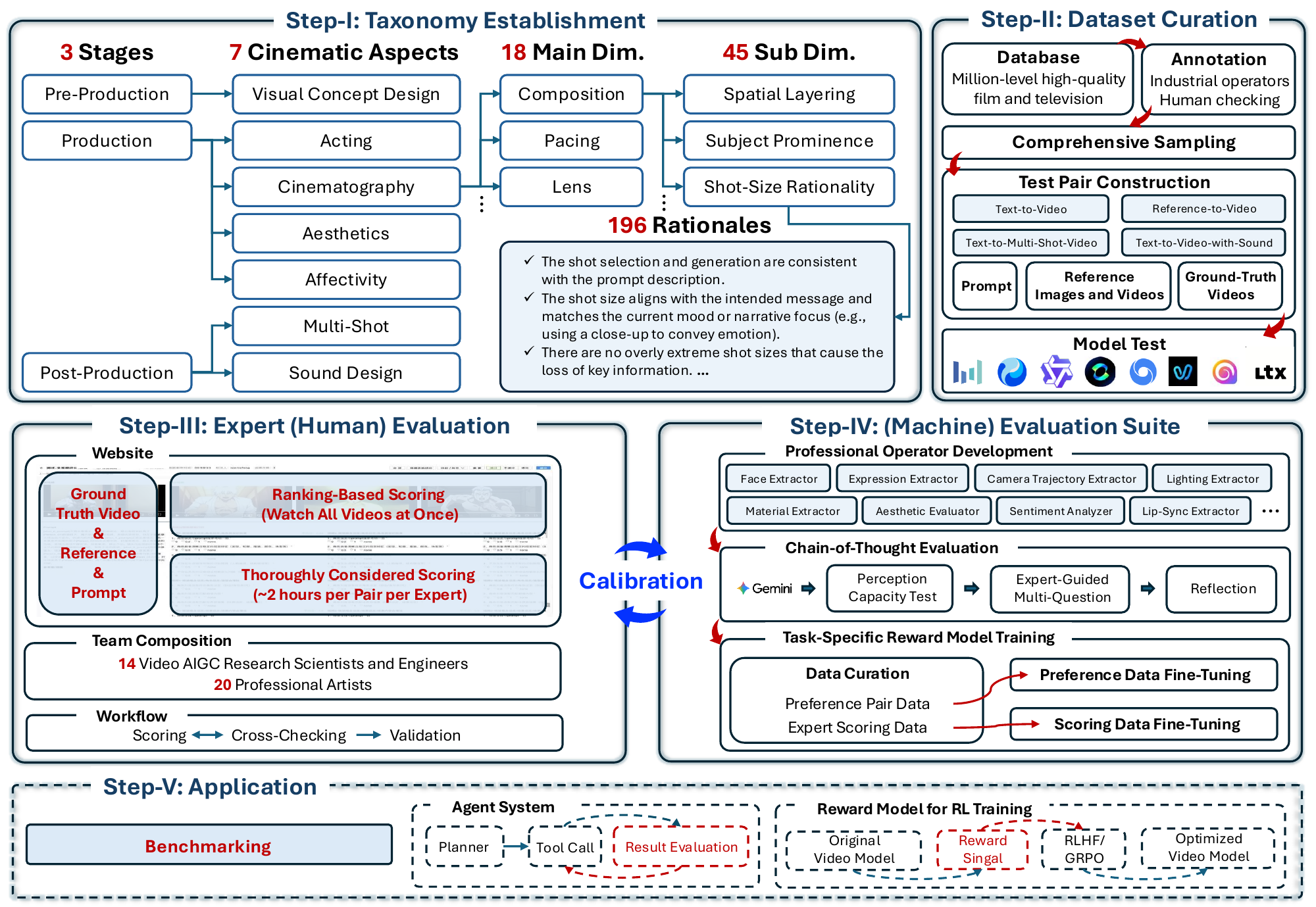}
  \vspace{-0.6cm}
  \caption{\textbf{Overview.} EvalVerse systematically digitizes subjective cinematic expertise into a computable, expert-calibrated evaluation framework through five steps. \textbf{(I) Taxonomy Establishment:} Decomposing the professional filmmaking workflow into 3 production stages, encompassing 7 cinematic aspects, 18 main dimensions, 45 sub-dimensions, and 196 granular rationales to structurally define cinematic ``goodness.'' \textbf{(II) Dataset Curation:} Constructing test pairs across full-modality video generation tasks (\textit{e.g.,} multi-shot, audio-visual) via comprehensive sampling from a million-scale professional database. \textbf{(III \& IV) Expert-Machine Calibration:} Bridging the historical divide between human aesthetic perception and algorithmic scoring. By synergizing specialized perception extractors with an expert-guided Chain-of-Thought process, we align Vision-Language Model reasoning with 34 professional experts. \textbf{(V) Versatile Applications:} Beyond static diagnostic benchmarking, EvalVerse serves as a fundamental infrastructure, showing promising potential to provide high-quality reward signals for Reinforcement Learning and act as an expert-level evaluator for autonomous video agent workflows.}
  \label{fig:teaser}
  % \vspace{-0.3cm}
\end{figure*}

\section{Introduction}

The rapid evolution of generative video foundation models~\cite{sora,hunyuanvideo,veo,seedance2,kling,wan} has propelled the field toward a new frontier of cinematic synthesis. Despite achieving remarkable pixel-level visual fidelity through massive Supervised Fine-Tuning (SFT)~\cite{vace}, a significant chasm remains between the raw output of these models and the demanding requirements of professional filmmaking. As SFT approaches a scalability bottleneck due to the scarcity of high-quality cinematic data, the field is transitioning toward Reinforcement Learning (RL) paradigms (\textit{e.g.,} RLHF~\cite{rlhf}, GRPO~\cite{dancegrpo}) and agentic workflows~\cite{Wu2025MovieAgent} to achieve precise control and complex narratives. In this new era, evaluation is no longer merely a passive leaderboard; it is becoming the critical bottleneck. Professional, reliable, fine-grained evaluation frameworks are therefore the essential prerequisite for providing high-quality reward signals and guiding the next generation of AI-aided cinematic evolution.

However, we observe a critical twofold gap in the current landscape of video generation evaluation. 
\textbf{(i) The ``Right'' vs. ``Good'' Objective Gap}: Existing benchmarks~\cite{evalcrafter,vbench,vbench++,vbench2,univbench} are predominantly stuck in the paradigm of evaluating ``whether it is right''—focusing merely on prompt-following capabilities and the basic presence of visual elements. They fundamentally fail to assess ``whether it is good,'' neglecting the nuanced aesthetic, physical, and cinematic qualities required for professional production. 
\textbf{(ii) Methodological and Credibility Gap}: The transition from evaluating ``rightness'' to ``goodness'' introduces a severe methodological bottleneck. Assessing cinematic quality inherently relies on domain-specific expert knowledge~\cite{vadb} and subjective nuances that previous automated metrics fundamentally fail to capture. Consequently, the field is trapped in an evaluation paradox: while professional human assessment is the gold standard, it is prohibitively expensive and unscalable; conversely, generic Vision-Language Models (VLMs)—the default automated alternative~\cite{datbench}—lack the professional rigor and domain-specific logic alignment.

To systematically address this twofold gap, we propose \textbf{EvalVerse} (Fig.~\ref{fig:teaser}), which takes a pragmatic first step in shifting the evaluation paradigm from generic visual scoring to a structured audit of professional filmmaking. Our framework directly resolves the aforementioned challenges through two corresponding technical contributions:

\begin{table*}[t]
\centering
\setlength{\tabcolsep}{14pt}
\caption{\textbf{Comparison of \textbf{EvalVerse} with existing video generation benchmarks.} Our framework is the first to achieve full-modality coverage across audio-sync and multi-shot sequencing, while introducing a pipeline-aware paradigm with high interpretability via expert-guided CoT.}
\label{tab:scope_comparison}
% \vspace{-0.3cm}
\resizebox{\textwidth}{!}{

\begin{tabular}{llccccccc}
\toprule
\multicolumn{2}{c}{\multirow{2}{*}{\textbf{Benchmark}}} & \multicolumn{4}{c}{\textbf{Task Modality Coverage}} & \multicolumn{3}{c}{\textbf{Evaluation Paradigm}} \\
\cmidrule(lr){3-6} \cmidrule(lr){7-9}
 && Text-to-Video & Reference-to-Video & Video with Sound & Multi-Shot & Pipeline-Aware & Expert-Guided & Interpretability \\
\midrule
EvalCrafter& \cite{evalcrafter} & \checkmark & $\times$ & $\times$ & $\times$ & $\times$ & $\times$ & Mid \\
VBench& \cite{vbench} & \checkmark & $\times$ & $\times$ & $\times$ & $\times$ & $\times$ & Low \\
VBench 2.0& \cite{vbench2} & \checkmark & $\times$ & $\times$ & $\times$ & $\times$ & $\times$ & Mid \\
VBench++& \cite{vbench++} & \checkmark & \checkmark & $\times$ & $\times$ & $\times$ & $\times$ & Mid \\
VADB& \cite{vadb} & $\times$ & $\times$ & $\times$ & $\times$ & $\times$ & \checkmark & High \\
CineTechBench& \cite{cinetechbench} & $\times$ &\checkmark  & $\times$ & $\times$ & $\times$ & \checkmark & Mid \\
Stable Cinemetrics &\cite{stable_cinemetrics} & \checkmark & $\times$ & $\times$ & $\times$ & Partial & \checkmark & Mid \\
UniVBench &\cite{univbench} & \checkmark & \checkmark & $\times$ & $\times$ & $\times$ & $\times$ & Mid \\
\midrule
\rowcolor[HTML]{E6F2FF} 
\textbf{EvalVerse} & (Ours)& \checkmark & \checkmark & \checkmark & \checkmark & \checkmark & \checkmark & \textbf{High (CoT)} \\
\bottomrule
\end{tabular}
}
\vspace{-0.3cm}
\end{table*}

\textbf{(i) Pipeline-Aware Cinematic Taxonomy}: To systematically define and measure ``goodness,'' we propose the first evaluation taxonomy that employs the professional filmmaking workflow as a structured diagnostic lens. Rather than assuming AI generation occurs in discrete steps, we audit the final generated video by mapping its complex multimodal elements back to three traditional production stages: pre-production (assessing foundational visual concept design), production (evaluating dynamic acting, cinematography, aesthetics, \& affectivity), and post-production (analyzing multi-shot \& sound design). This comprehensive framework captures the nuanced cinematic qualities neglected by previous benchmarks, enabling explainable diagnostic probing of specific model capabilities rather than just outputting a single holistic score.
    
\textbf{(ii) Expert-Calibrated Chain-of-Thought Evaluator}: To overcome the evaluation paradox and bridge the credibility gap of automated metrics, we introduce a massive human-in-the-loop calibration process involving professional domain experts (filmmakers and artists), algorithm scientists, and engineers. By repeatedly cross-calibrating human judgments with the actual perceptual and analytical boundaries of current state-of-the-art VLMs~\cite{gemini,Qwen3-VL}, we develop specialized evaluators that align their internal reasoning logic with professional critics. This pragmatic approach forces the evaluator to generate professional-grade Chain-of-Thought (CoT) rationales before scoring, successfully digitizing subjective, expert-level cinematic knowledge into scalable and interpretable machine metrics.

Furthermore, our comprehensive survey (Tab.~\ref{tab:scope_comparison}) reveals that existing video benchmarks~\cite{evalcrafter,vbench,vbench++,vbench2} significantly lag behind the rapid evolution of foundation models. They~\cite{cinetechbench,univbench,msvbench,muss} predominantly focus on silent, single-shot generation and construct test prompts by artificially permuting isolated cinematic elements~\cite{stable_cinemetrics}, failing to capture authentic cinematic distributions or provide reference videos for evaluation. To address these limitations, EvalVerse incorporates full-modality \& multi-shot narrative coverage. Supporting this evaluation is our ``Real-to-Gen'' data engine for test pair construction, which performs diversified, proportional sampling from real-world professional video datasets. Through hierarchical structural annotation and asset disentanglement, this engine generates high-fidelity test pairs with authentic reference videos, reflecting the true distribution of professional production and eliminating the stochastic bias inherent in existing prompt-based benchmarks.

In summary, EvalVerse treats video evaluation as a core scientific problem—the systematic digitization of subjective cinematic expertise—delivering two key contributions: \textbf{(i) Methodological Innovation:} By \textit{organizing} domain expertise into a pipeline-aware taxonomy, \textit{distilling} expert judgments into a curated dataset, and \textit{injecting} this knowledge into VLMs via human-machine calibration, we successfully translate abstract professional evaluation into scalable, expert-aligned CoT reasoning. \textbf{(ii) Comprehensive Coverage \& Alignment:} EvalVerse retains compatibility with ``rightness'' and ``goodness'' while pioneering the evaluation of complex multi-shot sequencing and audio-visual integration, achieving strong human-machine alignment across these advanced dimensions. Looking toward future generative video paradigms, EvalVerse goes beyond a leaderboard by providing trustworthy diagnostic signals, with strong potential to support high-quality reward modeling for Reinforcement Learning and to serve as an expert evaluator for agentic workflows.

\section{Related Work}

\subsection{Generative Video Foundation Model}

The landscape of generative video foundation models has rapidly advanced from early 3D U-Nets~\cite{svd} to scalable DiT~\cite{dit} and Flow Matching architectures~\cite{wan, hunyuanvideo}. Beyond architectural scaling, functional capabilities have shifted dramatically. Modern models have evolved from stochastic, silent generation to highly controllable, professional-grade production~\cite{cogvideox, luma}. Crucially, recent breakthroughs have successfully introduced end-to-end \textit{audio-visual integration}~\cite{sora2, ltx, kling, seedance2} and complex \textit{multi-shot narrative sequencing}~\cite{guo2025longcontexttuning, holocine, multishotmaster}. This paradigm shift from generating isolated clips to synthesizing cohesive, multimodal cinematic sequences demands entirely new evaluation frameworks.

\subsection{Benchmark for Video Generation}

\noindent
\textbf{Evolution of General Benchmarks: From Consistency to Faithfulness.} Early evaluation paradigms primarily relied on holistic metrics such as FVD~\cite{fvd} and CLIP-Score~\cite{clip}, which often failed to capture the nuances of temporal dynamics and semantic precision. The landscape shifted with the introduction of VBench~\cite{vbench}, which pioneered the decomposition of video quality into multiple hierarchical dimensions. This was further refined by VBench 2.0~\cite{vbench2}, which shifted the focus toward intrinsic faithfulness—addressing the misalignment between textual prompts and generated content in complex scenarios. Subsequent iterations~\cite{msvbench,muss,avgenbench} like VBench++~\cite{vbench++} expanded the suite's versatility to cover broader generative capabilities. Simultaneously, UniVBench~\cite{univbench} attempted to provide a unified evaluation for Video Foundation Models.

\noindent
\textbf{Professionalization: Cinematography and Aesthetics.} Recognizing that ``visual appeal'' in professional contexts is governed by cinematographic laws, a new wave of specialized benchmarks has emerged. Stable Cinemetrics~\cite{stable_cinemetrics} introduced a structured taxonomy for professional video, focusing on the precision of camera control and lighting. CineTechBench~\cite{cinetechbench} further narrowed this focus by evaluating a model's understanding and generation of specific cinematographic techniques. In parallel, the assessment of ``beauty'' has moved from subjective scoring to multidimensional auditing. VADB~\cite{vadb} established a large-scale database with professional-grade annotations for video aesthetics. These works highlight a clear trend: the evaluation for video generation is moving beyond basic prompt-following toward the mastery of the visual language of cinema.

% \subsection{EvalVerse: Differentiating from Existing Benchmarks}

% Existing benchmarks often evaluate in isolation, either overemphasizing low-level visual consistency or focusing exclusively on niche cinematographic nodes. EvalVerse distinguishes itself by striking a strategic balance across three critical pillars: \textbf{(i) Generative Capacity:} Expanding beyond silent, single-shot generation to cover the latest functional frontiers, including reference-based and audio-visual synthesis. \textbf{(ii) Industrial Pipeline:} Reverse-engineering the evaluation of end-to-end videos through the diagnostic lens of pre-production, production, and post-production stages. \textbf{(iii) VLM Perception:} Explicitly acknowledging the sensing limitations of current Vision-Language Models to design metrics that are both professionally rigorous and practically computable. Unlike previous works relying on stochastic prompt sampling, EvalVerse aligns expert filmmaking logic with AI-native COT workflows, delivering the first comprehensive and actionable audit of professional cinematic generation.

\section{Taxonomy}

The core of EvalVerse is a hierarchical, pipeline-aware taxonomy designed to bridge the gap between AI video synthesis and professional filmmaking standards. Recognizing that modern foundation models typically synthesize videos in an end-to-end manner, we do not assume a multi-step generation process. Instead, we employ the traditional filmmaking workflow as a powerful diagnostic lens. Rather than treating the final generated video as a flat collection of visual attributes, our taxonomy reverse-engineers the assessment by mapping the complex multi-modal elements of the output onto three distinct conceptual stages: Pre-Production, Production, and Post-Production.

\begin{figure*}
    \centering
    \includegraphics[width=\linewidth]{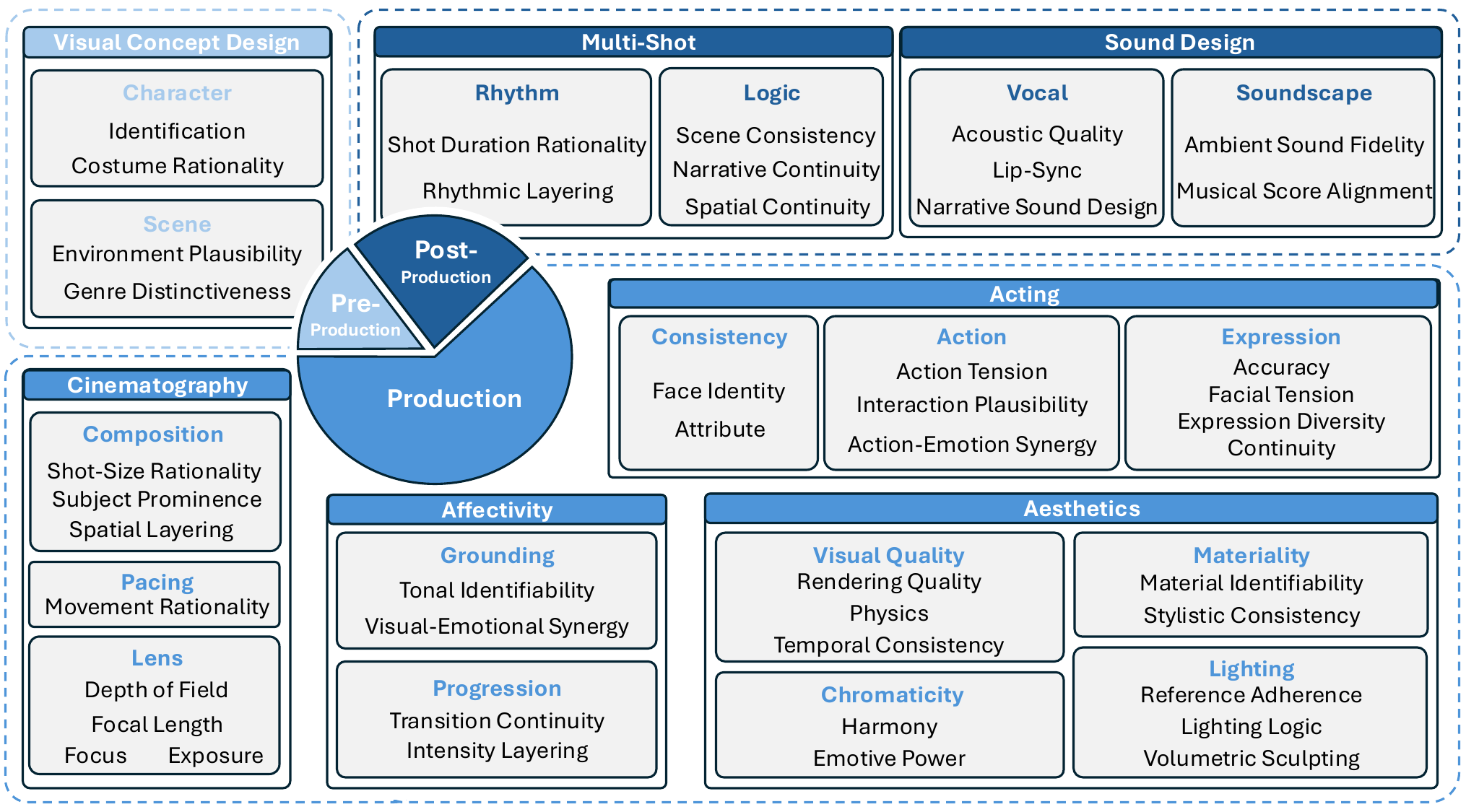}
    \vspace{-0.6cm}
    \caption{\textbf{Pipeline-aware evaluation taxonomy.} We propose a comprehensive taxonomy that mirrors the professional cinematic workflow.}
    \label{fig:taxonomy}
    % \vspace{-0.3cm}
\end{figure*}

\begin{table*}[t]
\centering
\small % 这里可以修改字体大小，可选 \footnotesize, \scriptsize, \small
\caption{\textbf{Comparison of evaluation dimensions across the full production pipeline.}}
\label{tab:taxonomy_comparison}
% \vspace{-0.3cm}
\resizebox{\textwidth}{!}{
\begin{tabular}{llccccccccccccc}
\toprule
\multicolumn{2}{c}{\multirow{2}{*}{\textbf{Benchmark}}} & \textbf{Pre-Prod.} & \textbf{Prod: Acting} & \multicolumn{3}{c}{\textbf{Prod: Cinematography}} & \multicolumn{4}{c}{\textbf{Prod: Aesthetics}} & \multicolumn{2}{c}{\textbf{Prod: Affectivity}} & \multicolumn{2}{c}{\textbf{Post-Prod.}} \\
\cmidrule(lr){3-3} \cmidrule(lr){4-4} \cmidrule(lr){5-7} \cmidrule(lr){8-11} \cmidrule(lr){12-13} \cmidrule(lr){14-15}
& & Design & Acting & Composition & Lens & Pacing & Vis. Quality & Chromaticity & Materiality & Lighting & Grounding & Progression & Multi-Shot & Sound \\
\midrule
EvalCrafter& \cite{evalcrafter} & $\times$ & $\times$ & $\times$ & $\times$ & $\times$ & \checkmark & $\times$ & $\times$ & $\times$ & $\times$ & $\times$ & $\times$ & $\times$ \\
VBench& \cite{vbench} & $\times$ & Partial & $\times$ & $\times$ & $\times$ & \checkmark & $\times$ & $\times$ & $\times$ & $\times$ & $\times$ & $\times$ & $\times$ \\
VBench 2.0& \cite{vbench2} & $\times$ & Partial & $\times$ & $\times$ & $\times$ & \checkmark & $\times$ & $\times$ & $\times$ & $\times$ & $\times$ & $\times$ & $\times$ \\
VBench++& \cite{vbench++} & $\times$ & Partial & $\times$ & $\times$ & $\times$ & \checkmark & $\times$ & $\times$ & $\times$ & $\times$ & $\times$ & $\times$ & $\times$ \\
VADB& \cite{vadb} & Partial & Partial & Partial & Partial & $\times$ & \checkmark & \checkmark & $\times$ & Partial & Partial & $\times$ & $\times$ & $\times$ \\
CineTechBench& \cite{cinetechbench} & $\times$ & $\times$ & \checkmark & \checkmark & \checkmark & $\times$ & \checkmark & $\times$ & \checkmark & $\times$ & $\times$ & $\times$ & $\times$ \\
Stable Cinemetrics& \cite{stable_cinemetrics} & Partial & \checkmark & Partial & \checkmark & Partial & $\times$ & $\times$ & $\times$ & Partial & Partial & $\times$ & $\times$& $\times$ \\
UniVBench& \cite{univbench} & $\times$ & Partial & Partial & \checkmark & Partial & \checkmark & Partial & $\times$ & Partial & $\times$ & $\times$ & $\times$ & $\times$ \\
\midrule
\rowcolor[HTML]{E6F2FF} 
\textbf{EvalVerse}&  (Ours) & \checkmark & \checkmark & \checkmark & \checkmark & \checkmark & \checkmark & \checkmark & \checkmark & \checkmark & \checkmark & \checkmark & \checkmark & \checkmark \\
\bottomrule
\end{tabular}
}
\vspace{-0.3cm}
\end{table*}

\subsection{Pre-Production}
This stage evaluates the foundational ``Visual Development'' and asset design logic before dynamic synthesis occurs. It ensures that the generated assets possess clear identifiability and logical consistency.

\subsubsection{Visual Concept Design}
\noindent
As the cornerstone of directing and art design, this dimension audits the conceptual integrity of characters and environments, ensuring they align with the intended worldview and narrative settings.

\noindent
\textbf{Character.} This dimension audits the foundational asset integrity of the subject. It encompasses \textit{Identifiability}, which requires clear, recognizable visual anchors (\textit{e.g.,} unique facial structures, body types, and silhouettes) that distinguish the character from others without identity morphing (such as unintended changes in face or clothing). It also includes \textit{Costume Rationality}, which evaluates whether the character's attire and styling logically match their intended concept (profession, identity, era), the specific scene context, and the overarching worldview.

\noindent
\textbf{Scene.} This focuses on the world-building logic of the environment. It includes \textit{Environment Plausibility}, auditing whether the spatial arrangement of objects follows physical laws (\textit{e.g.,} gravity, collisions, support) and spatial logic (perspective, scale, relations), penalizing AI hallucinations like floating objects or clipping. Furthermore, \textit{Genre Distinctiveness} measures the purity of the artistic style, ensuring that the visual language (whether realism, animation, or cyberpunk) exhibits clear, characteristic signatures in lighting, materials, and colors, without inappropriate stylistic mixing (\textit{e.g.,} blending 2D and 3D elements illogically).

\subsection{Production}
This stage evaluates the execution of the ``virtual shoot.'' It comprehensively assesses how the subject performs, how the camera captures the scene, the overall visual aesthetics, and the emotional atmosphere generated.

\subsubsection{Acting}
\noindent
This dimension evaluates the subject's presentation, focusing on the dynamic consistency, physical kinetic power, and psychological nuance of the character's performance.

\noindent
\textbf{Consistency.} This ensures the stability of character assets during movement. It includes \textit{Face Identity}, requiring facial features to remain consistent across varying angles without morphing or AI-induced structural changes during motion. It also covers \textit{Attribute} consistency, ensuring that hair length/color, clothing style/material, and accessories remain stable without sudden flickering, disappearing, or unintended transformations.

\noindent
\textbf{Action.} This evaluates the kinetic power, narrative intent, and physical interactions of movement. It covers \textit{Action Tension}, ensuring movements follow physical logic (avoiding mechanical or weightless motions) and possess natural kinetic force without biological impossibilities (\textit{e.g.,} bone breaking). It also includes \textit{Action-Emotion Synergy}, assessing whether the physicality reflects the character's internal state (\textit{e.g.,} anger driving forceful actions, joy driving lightness) and effectively drives the emotional narrative. Furthermore, it evaluates \textit{Interaction Plausibility}, ensuring that interactions align with prompt descriptions, demonstrate clear contact and basic force logic, and avoid generation errors like clipping or incorrect positioning, while maintaining logical displacement, movement, and deformation of the interacted objects.

\noindent
\textbf{Expression.} This assesses the nuance of the character's facial performance. Metrics include \textit{Accuracy} (matching the text prompt and contextual logic without contradictory expressions), \textit{Facial Tension} (natural muscle contractions and micro-expressions, avoiding over-exaggeration or stiffness), \textit{Expression Diversity} (providing layered, rich emotional changes rather than a monotonous single expression), and \textit{Continuity} (ensuring smooth, biologically plausible emotional transitions without abrupt jumps).

\subsubsection{Cinematography}
\noindent
This dimension evaluates the ``virtual camera'' language and visual storytelling, auditing how the framing, optical properties, and camera movements serve the narrative.

\noindent
\textbf{Composition.} This evaluates the framing logic. It includes \textit{Shot-Size Rationality} (appropriateness of close-ups vs. wide shots for the narrative, avoiding awkward framing like cutting off heads), \textit{Subject Prominence} (ensuring the main subject is visually salient, not obscured by lighting or messy backgrounds, and effectively guides the viewer's eye), and \textit{Spatial Layering} (establishing clear foreground, midground, and background separation, utilizing light and shadow for depth, and maintaining spatial continuity during movement).

\noindent
\textbf{Lens.} This audits the physical validity of the camera's optical settings. It encompasses \textit{Depth of Field} (clear focal planes, natural bokeh gradients, and logical depth changes during movement without edge artifacts), \textit{Focal Length} (adhering to the perspective logic of wide, standard, or telephoto lenses based on spatial constraints), \textit{Focus} (clear focus points, logical focus shifts, and tracking, avoiding sudden focus jumps or blurring of key areas), and \textit{Exposure} (maintaining appropriate dynamic range, matching the scene's lighting context, and avoiding AI-induced exposure flickering).

\noindent
\textbf{Pacing.} This evaluates the temporal dynamics of camera movement. It focuses on \textit{Movement Rationality}, ensuring that camera trajectories (pan, tilt, push, pull) serve a clear narrative purpose, possess appropriate speed and natural kinetic inertia, and are free from unintended AI shaking or aimless drifting.

\subsubsection{Aesthetics}
\noindent
This dimension focuses on the technical fidelity and artistic rendering of the video, encompassing visual quality, color grading, physical materiality, and lighting design.

\noindent
\textbf{Visual Quality.} This focuses on the foundational render fidelity, physical accuracy, and temporal stability of the generated content. \textit{Rendering Quality} ensures sufficient clarity and high resolving power for distinguishable details. It penalizes visual degradations such as noise, grain, compression artifacts, and edge anomalies (\textit{e.g.,} aliasing or ghosting). Furthermore, it requires rich textural details, avoiding overly smooth or ``plastic'' appearances, and strictly prohibits generative artifacts like distortions or repetitive textures. \textit{Physics} evaluates adherence to real-world physical principles, ensuring logical physical morphology and structural details (avoiding shadow, reflection, or structural errors). It ensures objects obey basic physical laws (\textit{e.g.,} gravity, inertia, and material properties), demonstrate plausible force and interaction feedback without weightless or floating effects, and follow rational movement and displacement paths. Finally, \textit{Temporal Consistency} assesses stability across continuous frames, penalizing fluctuations in clarity, detail flickering or repainting, edge jittering, brightness or color flashes, and sudden local quality degradation (\textit{e.g.,} localized collapse).

\noindent
\textbf{Chromaticity.} This audits the artistic use of color. It includes \textit{Harmony} (balanced color grading, unified tones, and absence of abrupt/messy colors) and \textit{Emotive Power} (how the palette amplifies the intended mood, changes dynamically with the narrative, and utilizes color contrast for visual emphasis).

\noindent
\textbf{Materiality.} This evaluates surface realism through \textit{Material Identifiability} (accurate optical properties like reflection, roughness, and transparency to distinguish metal, skin, fabric, or glass, avoiding plastic-looking skin) and \textit{Stylistic Consistency} (unified shader language across assets that matches the overall lighting and artistic style).

\noindent
\textbf{Lighting.} This audits the illumination logic. It includes \textit{Lighting Logic} (matching the prompt's specified directional/ambient light, time of day, and color temperature, clear light sources, consistent shadow directions and intensities, and absence of unexplained light leaks), and \textit{Volumetric Sculpting} (how light defines 3D form, spatial depth, and maintains volume dynamically during movement).

\subsubsection{Affectivity}
\noindent
This dimension evaluates the emotional resonance and atmospheric setup of the video, ensuring that the visual elements collectively generate a compelling and continuous emotional experience.

\noindent
\textbf{Grounding.} This assesses the initial atmospheric setup. It includes \textit{Tonal Identifiability} (establishing a clear emotional baseline, such as neutral, tense, warm, or depressing, that fits the narrative context) and \textit{Visual-Emotional Synergy} (ensuring color, light, composition, and lens choices collectively express the emotion, avoiding conflicts between visual presentation and the intended mood).

\noindent
\textbf{Progression.} This evaluates the emotional arc over time. It audits \textit{Transition Continuity} (smoothness of emotional shifts without causeless, abrupt jumps) and \textit{Intensity Layering} (the presence of emotional buildup or decrescendo, utilizing visual techniques to enhance emotional peaks, avoiding overly flat or excessively explosive expressions, and adhering to a rhythmic structure of setup, development, and climax).

\subsection{Post-Production}
The final stage evaluates the assembly of shots and the multimodal integration, focusing on multi-shot logic and audio-visual synchronization. While these evaluation dimensions could theoretically extend to traditional editing and dubbing, assessing complex post-processing interventions remains highly challenging. Therefore, our current scope strictly focuses on natively generated multi-shot sequences and synthesized audio.

\subsubsection{Multi-Shot}
\noindent
This dimension evaluates the sequential logic and temporal rhythm between multiple shots, ensuring narrative flow and spatial coherence.

\noindent
\textbf{Logic.} This evaluates sequential continuity. Metrics include \textit{Scene Consistency} (stable environments, props, lighting, weather, and character makeup across cuts without AI generation errors), \textit{Narrative Continuity} (logical cause-and-effect action sequencing and subject state maintenance), and \textit{Spatial Continuity} (adherence to character positioning, the 180-degree rule, and spatial orientation).

\noindent
\textbf{Rhythm.} This audits the temporal heartbeat of the edit. It includes \textit{Shot Duration Rationality} (providing sufficient time for information consumption, matching the emotional tension and audio rhythm) and \textit{Rhythmic Layering} (varying cutting tempos, combining dynamic and static shots, and aligning the rhythm with the narrative arc from setup to climax).

\subsubsection{Sound Design}
\noindent
This dimension evaluates the relationship between sound and image, auditing the integration of human voices, ambient sounds, and musical scores.

\noindent
\textbf{Vocal.} This evaluates human voice integration. It includes \textit{Acoustic Quality} (matching the character's age, gender, and personality, ensuring technical purity without mechanical noise, and maintaining spatial reverb consistency) and \textit{Lip-Sync} (temporal alignment between phonemes and mouth shapes, matching mouth opening with volume, and aligning vocal emotion with facial expression), and \textit{Narrative Sound Design} (using off-screen audio cues to convey emotion, guide audience attention, and expand the narrative space).

\noindent
\textbf{Soundscape.} This focuses on the immersive sonic environment. It audits \textit{Ambient Sound Fidelity} (realism of Foley, spatial depth, and matching the on-screen environment/weather) and \textit{Musical Score Alignment} (synchronizing musical tone and rhythm with visual cuts and emotional beats without overpowering dialogue).

\begin{figure*}
    \centering
    \includegraphics[width=\linewidth]{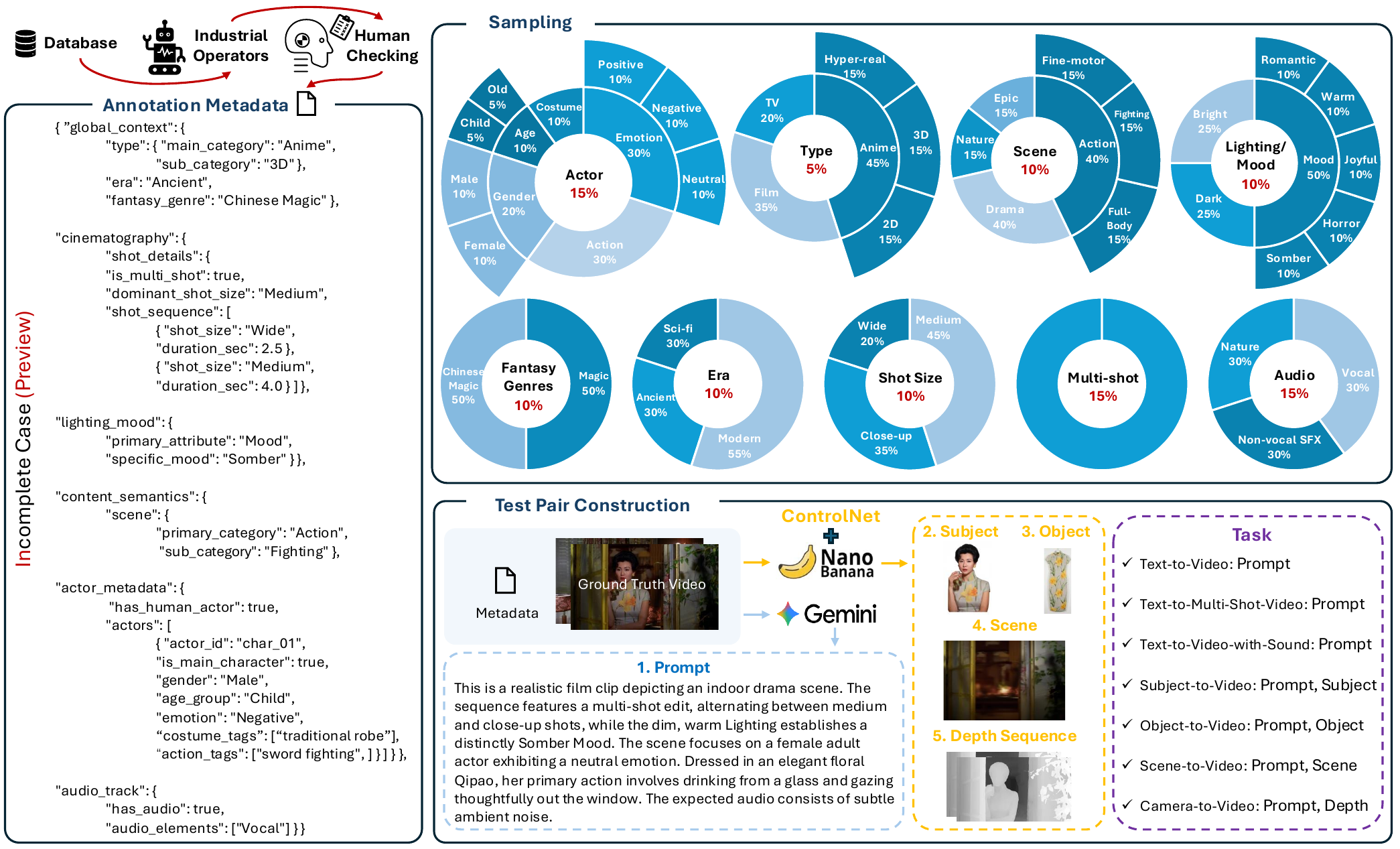}
    \vspace{-0.6cm}
    \caption{\textbf{Comprehensive pipeline for dataset annotation, sampling, and test pair construction.} \textbf{(Left)} The annotation pipeline, yielding structured JSON metadata via industrial operators and human verification. \textbf{(Top Right)} Proportional distributions ensuring balanced and comprehensive data sampling. \textbf{(Bottom)} Test pair construction generating multi-modal inputs for diverse downstream generation tasks.}
    \label{fig:data_sampling}
\end{figure*}

\section{Dataset Curation: Test Pair Construction}

To capture professional filmmaking complexities in the EvalVerse benchmark, our data engine (Fig.~\ref{fig:data_sampling}) transforms raw cinematic videos into ``Real-to-Gen'' test pairs via structured annotation, strategic sampling, and test pair construction.

% We begin with a high-quality video database comprising diverse professional cinema and stylized animation. To transform these unstructured pixels into evaluable metadata, we employ a multi-modal perception suite. This pipeline performs structured annotation on every clip, extracting a rich set of tags that comprehensively cover all elements defined in our evaluation taxonomy, spanning technical camera parameters, character attributes, and environmental contexts. To guarantee utmost reliability, all extracted metadata are processed by industrial-grade annotation operators and subsequently subjected to rigorous manual verification. Consequently, these highly accurate labels serve as the robust ``ground truth'' metadata for our downstream sampling and prompt generation.

\noindent \textbf{Annotation.} Starting with a diverse database of professional films and animations, we employ a multi-modal perception suite to extract structured metadata covering our entire evaluation taxonomy (\textit{e.g.,} camera parameters, character attributes, and environments). Following industrial-grade processing and rigorous manual verification, these highly accurate labels serve as robust ground-truth metadata for downstream sampling and prompt generation.

\noindent \textbf{Sampling.} To ensure the benchmark is both comprehensive and industry-representative, we perform diversified sampling from the annotated database. Rather than a stochastic selection, we adopt a proportional sampling strategy across nine core cinematic dimensions to maintain a balanced distribution.

\noindent
 \textbf{Construction.} This stage involves the construction of multi-modal test pairs tailored for different generation tasks. We utilize a Gemini 3.1 Pro~\cite{gemini} to ingest the structured metadata and raw video captions, synthesizing professional-grade test prompts that reflect cinematic terminology. For reference-based tasks (\textit{e.g.,} subject-driven generation), we extract keyframes from the source videos and employ Nano Banana Pro~\cite{nanobanana} to generate high-fidelity reference images. For depth reference, we adopt ControlNet-tuned~\cite{controlnet} model to generate depth sequences.

\section{Benchmark: Expert Evaluation Results}

\subsection{Benchmarking Settings}

\noindent
\textbf{Human Evaluation Protocol.} To guarantee both cinematic aesthetics and algorithmic fidelity, our evaluation is conducted by a multi-disciplinary team (filmmakers, AIGC scientists, and engineers) through a strict three-stage pipeline: (i) Discriminative Annotation: Annotators perform side-by-side comparisons of the prompt, Ground Truth video, and model outputs. Crucially, to yield strong preference signals, they must assign strict discriminative rankings across all predefined dimensions. (ii) Quality Assurance: Senior film industry professionals conduct item-by-item reviews of the initial annotations, assigning Pass/Fail labels based on cinematic validity and consistency. (iii) Final Audit: Experts oversee the ultimate verification, resolving anomalies and eliminating systemic bias to finalize the ground-truth human preference dataset.

\noindent
\textbf{Video Generation Model Selection.} (i) Closed-Source Models: We include Kling-v3-Omni~\cite{kling}, Seedance 2.0~\cite{seedance2}, Happy Horse 1.0~\cite{happyhorse}, Vidu-Q2-Pro~\cite{vidu}, and Hailuo 2.3~\cite{hailuo}. (ii) Open-Source Models: We evaluate Hunyuan 1.5 (8.3B)~\cite{hunyuanvideo}, LTX2 (19B)~\cite{ltx}, and Wan2.2 (14B)~\cite{wan}. (iii) Multi-Shot and Audio-Visual Models: We specifically select models that push the boundaries of multimodal cinematic synthesis. On one hand, models like HoloCine (14B)~\cite{holocine} and MultiShotMaster (14B)~\cite{multishotmaster} are at the frontier of multi-shot narrative sequencing. On the other hand, models such as Kling-v3-Omni, Seedance 2.0, Happy Horse 1.0, and LTX2 feature native sound design capabilities.

 \begin{figure*}
    \centering
    \includegraphics[width=\linewidth]{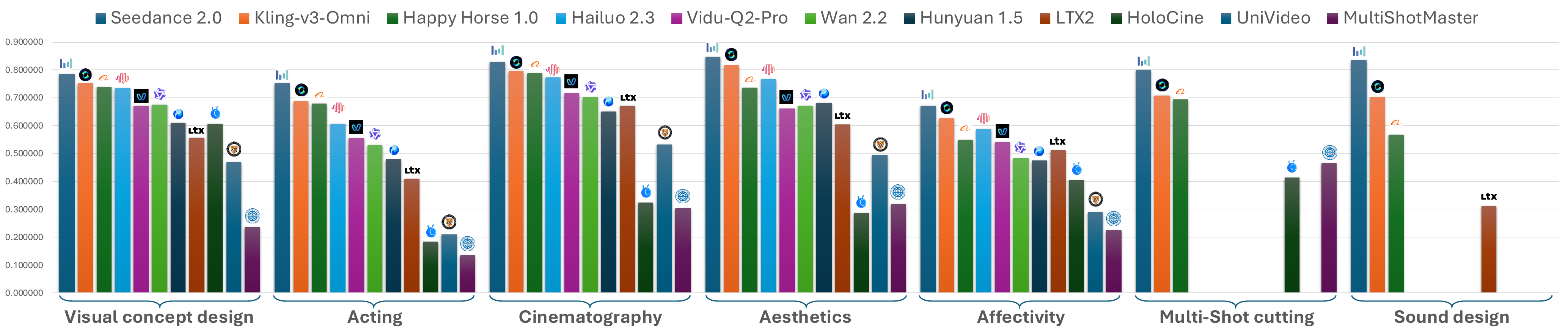}
    \vspace{-0.6cm}
    \caption{\textbf{Overall performance comparison of evaluated video generation models.}}
    \label{fig:main_dim_overview}
\end{figure*}

\begin{figure*}
    \centering
    \includegraphics[width=\linewidth]{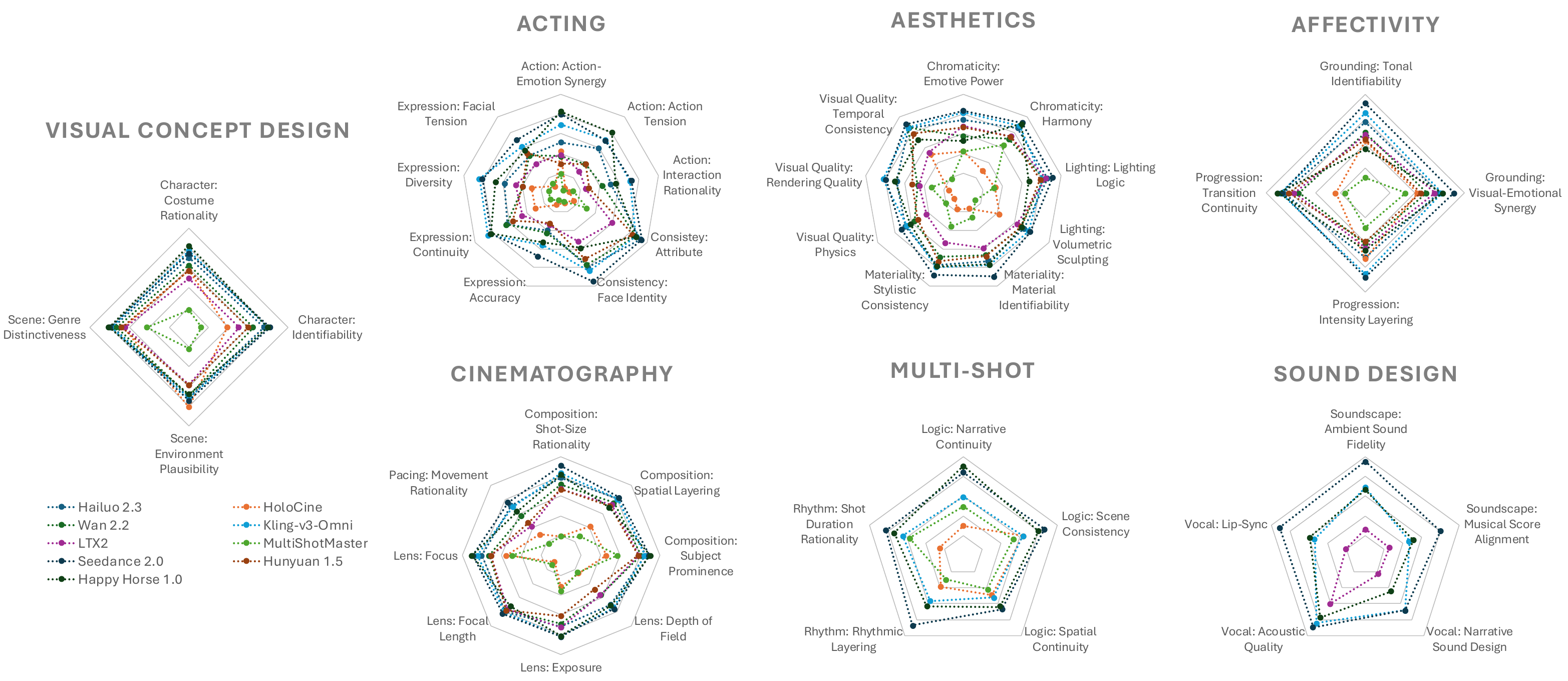}
    \vspace{-0.6cm}
    \caption{\textbf{Fine-grained performance comparison of evaluated models in the Text-to-Video (T2V) setting.}}
    \label{fig:human_evaluation_t2v}
\end{figure*}

 \begin{figure*}
    \centering
    \includegraphics[width=\linewidth]{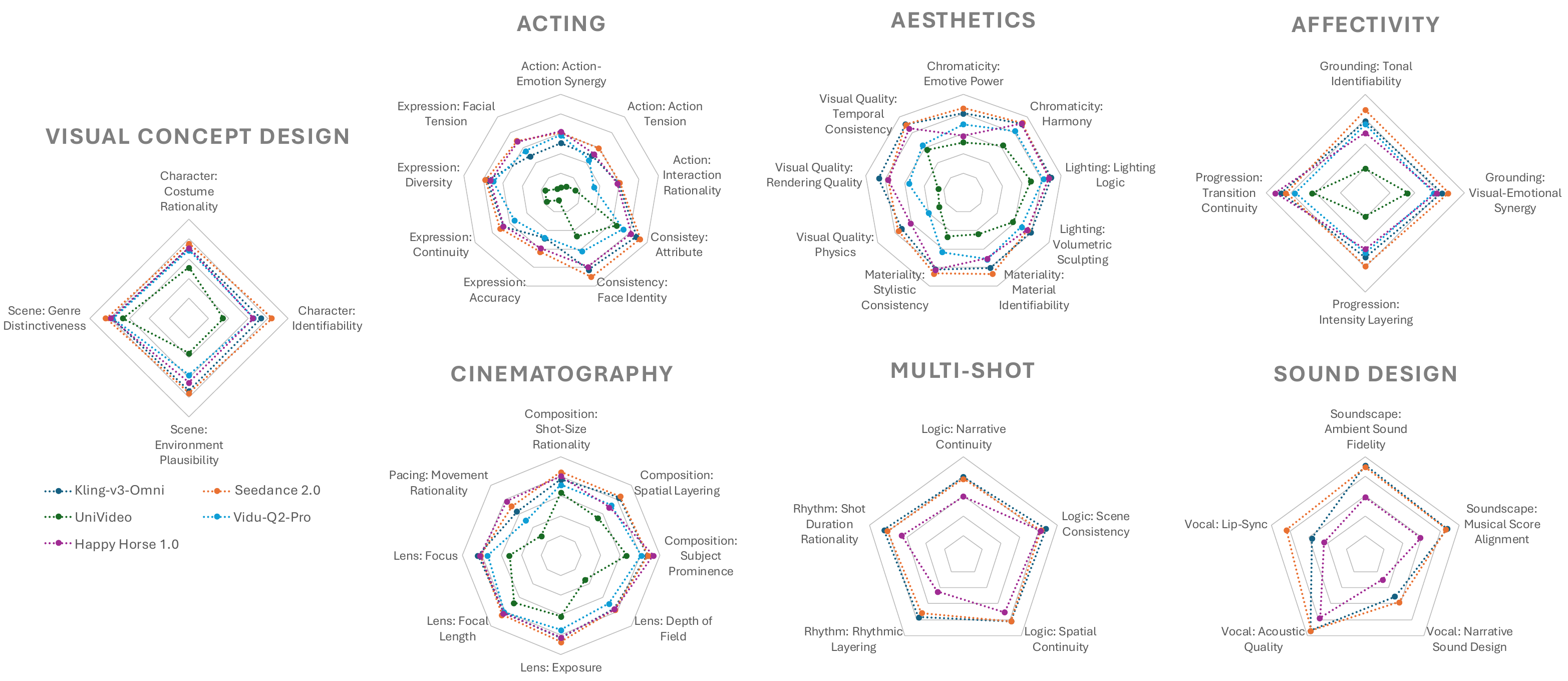}
    \vspace{-0.6cm}
    \caption{\textbf{Fine-grained performance comparison of evaluated models in the Reference-to-Video (R2V) setting.}}
    \label{fig:human_evaluation_r2v}
\end{figure*}

\subsection{Benchmarking Analysis}

\noindent
\textbf{Overall.} As shown in Fig.~\ref{fig:main_dim_overview}, the evaluated models exhibit a clear hierarchical distribution. Seedance 2.0 achieves the best comprehensive performance, demonstrating consistently strong results in every way. Kling-v3-Omni and Happy Horse 1.0 constitute the next leading group: the former shows strong and stable performance in aesthetics, cinematography, visual concept preservation, and sound-related dimensions, while the latter performs particularly well in cinematography, visual concept design, acting, and multi-shot organization. In comparison, Hailuo 2.3 and Vidu-Q2-Pro form a competitive middle tier, with strengths mainly concentrated in cinematography, aesthetics, and visual concept design, but relatively weaker performance in affectivity and sound-related dimensions. Wan2.2, Hunyuan 1.5, and LTX2 show moderate overall capability, with advantages mainly in visual and camera-related criteria, whereas HoloCine, UniVideo, and MultiShotMaster present more uneven or specialized performance profiles.

% Overall, the results suggest that the leading models are distinguished not only by visual quality, but also by their ability to maintain balanced cinematic competence across semantic, temporal, aesthetic, and audio-visual dimensions.

\noindent
\textbf{Text-to-Video.} As shown in Fig.~\ref{fig:human_evaluation_t2v}, Seedance 2.0 remains the strongest overall model, achieving the highest average score and consistently ranking near the top across most fine-grained criteria. Its performance is particularly strong in soundscape fidelity, identity preservation, attribute consistency, visual quality, and camera control, indicating robust semantic preservation and balanced perceptual quality. Kling-v3-Omni also maintains a leading position, with strong results in visual concept design, aesthetics, cinematography, and attribute consistency, though its sound-design-related scores are relatively less dominant. Happy Horse 1.0 emerges as a highly competitive newly evaluated model, showing strong performance in chromatic harmony, narrative continuity, subject prominence, visual concept design, and cinematography, while remaining weaker in tonal grounding and narrative sound design. The remaining models show more specialized or uneven profiles. Hailuo 2.3 and Wan2.2 form a competitive middle tier: Hailuo 2.3 performs well in lighting logic, temporal consistency, and camera-related criteria, while Wan2.2 shows strengths in visual consistency and concept preservation but is weaker in action and expression-related dimensions. Hunyuan 1.5 demonstrates moderate visual generation quality, especially in attribute consistency and lighting-related criteria, but its action interaction and expression dimensions are less competitive. LTX2 performs reasonably in lighting and lens-related criteria, yet shows limitations in expression, rhythm, and audio-visual synchronization. HoloCine presents localized strengths in scene plausibility and multi-shot consistency, but lower acting and rendering-related scores limit its overall quality. MultiShotMaster performs relatively well in shot-duration rationality while lagging behind in broader visual concept design, acting, aesthetics, and sound dimensions.

\noindent
\textbf{Reference-to-Video.} As shown in Fig.~\ref{fig:human_evaluation_r2v}, Seedance 2.0, Kling-v3-Omni, and Happy Horse 1.0 form the leading group. Seedance 2.0 achieves the highest overall score and maintains strong performance across most evaluated tertiary dimensions, with particularly strong results in vocal acoustic quality, chromatic harmony, attribute consistency, visual concept preservation, and camera-related criteria. Kling-v3-Omni ranks second overall and remains highly competitive, performing strongly in vocal acoustic quality, chromatic harmony, ambient sound fidelity, lens focus, and multi-shot scene consistency, although its action tension and expression-related scores are relatively less dominant. Happy Horse 1.0 also shows a stable and competitive profile, with strengths in subject prominence, chromatic harmony, lighting logic, visual concept design, and cinematography, but weaker results in multi-shot rhythm, lip-sync, and narrative sound design. Vidu-Q2-Pro forms a moderate performance tier. It performs well in lighting logic, chromatic harmony, focal-length control, costume rationality, and subject prominence, but shows weaker results in action interaction, physical realism, movement pacing, and narrative sound design. UniVideo exhibits the most uneven R2V performance: it retains reasonable scores in static visual attributes such as lighting logic, focal length, genre distinctiveness, spatial continuity, and shot-duration rationality, but falls behind on acting, expression dynamics, action-emotion synergy, affective progression, and sound-related dimensions. 

% Overall, the R2V comparison suggests that leading models are increasingly differentiated by their ability to preserve reference semantics while maintaining stable cinematography, visual quality, temporal coherence, and audio-visual consistency.

% The ultimate objective of our evaluation framework is to mathematically model the nuanced annotations of human cinematic experts, denoted as $\mathcal{H}$. However, empirical observations reveal a significant gap: relying solely on existing objective operators or the zero-shot perception and reasoning capabilities of VLMs is insufficient to fully fit the human expert distribution $\mathcal{H}$. To bridge this perception-reasoning gap, we conducted multiple rounds of iterative calibration against human evaluation. This process culminated in the finalization of our comprehensive cinematic taxonomy and the development of a hybrid machine evaluation paradigm. This paradigm synergizes professional operators, expert-guided multi-question, and a task-specific VLM fine-tuning to enable a reliable CoT reasoning. Formally, given a generated video sequence $V \in \mathbb{R}^{T \times H \times W \times C}$, its accompanying audio track $A$, the driving textual prompt $p$, and the reference $r$ , our goal is to derive a multi-dimensional cinematic score vector $\mathbf{S} \in \mathbb{R}^D$ such that $\mathbf{S} \approx \mathcal{H}(V, A, p, r)$. To achieve this, our pipeline is built upon three foundational modules that dynamically interact:

\section{Machine Evaluation Suite}

The ultimate objective of our framework is to mathematically model human expert annotations, denoted as $\mathcal{H}$. Formally, given a generated video $V \in \mathbb{R}^{T \times H \times W \times C}$, audio $A$, text prompt $p$, and reference $r$, our framework computes a multi-dimensional cinematic score vector $\mathbf{S} \in \mathbb{R}^D$ to approximate expert judgment: $\mathbf{S} \approx \mathcal{H}(V, A, p, r)$. To achieve this, we design a systematic evaluation pipeline (Sec.~\ref{sec: machine pipeline}) powered by a fine-tuned VLM (Sec.~\ref{sec: vlm fine_tuning}).

\subsection{Expert-Calibrated Evaluation Pipeline}
\label{sec: machine pipeline}
During the inference phase, our evaluation pipeline operates through a two-step mechanism: extracting deterministic evidence via specialized operators, followed by step-by-step reasoning using the fine-tuned VLM guided by expert multi-questioning.

\subsubsection{Professional Operator Extraction (Perception Prior)} VLMs inherently struggle with fine-grained temporal tracking and low-level perception. To mitigate hallucinations and provide reliable contextual priors, we first deploy a suite of specialized operators $\Phi = \{\phi_1, \dots, \phi_K\}$ to extract deterministic, objective evidence $E_{\text{prof}}$:
\begin{equation}
    E_{\text{prof}} = \bigcup_{k=1}^{K} \phi_k(V, A, p, r),
\end{equation}
where $\Phi$ includes operators such as DINO~\cite{oquab2023dinov2} and InsightFace~\cite{arcface} for cross-frame identity tracking, YOLO~\cite{khanam2024yolov11} for semantic anchoring, SyncNet~\cite{syncnet} for audio-visual synchronization, and Whisper~\cite{whisper} for speech emotion recognition.

\subsubsection{Expert-Guided CoT Reasoning \& Scoring} Equipped with the perception prior $E_{\text{prof}}$, the core evaluation is performed by our fine-tuned VLM, denoted as $\mathcal{M}_{\theta^*}$ (training details in Sec.~\ref{sec: vlm fine_tuning}). Let $X = (A, p, r, E_{\text{prof}}, \mathcal{Q})$ denote the comprehensive multi-modal context, where $\mathcal{Q}$ represents expert-designed multi-questions for a specific cinematic dimension. Rather than outputting a direct score blindly, $\mathcal{M}_{\theta^*}$ performs step-by-step reasoning, generating a detailed CoT. 

Crucially, this reasoning phase incorporates a \textit{Self-Reflection} mechanism within the CoT, forcing the VLM to take a step back and re-examine whether its judgments and reasoning have been subject to any hallucinations. Furthermore, we introduce a \textit{Context-Aware Gating} mechanism, represented by an indicator function $\mathbb{I}_{gate}(p, C) \in \{0, 1\}$, which dynamically bypasses specific metrics (\textit{e.g.,} strong expressions) if the narrative context $C$ does not warrant them. The final score $S_d$ for dimension $d$ is computed as:
\begin{equation}
    S_d = \mathcal{M}_{\theta^*}(V, X) \cdot \mathbb{I}_{gate}(p, C).
\end{equation}

\subsection{Two-Stage VLM Fine-Tuning for Human Alignment}
\label{sec: vlm fine_tuning}
To equip the foundational VLM $\mathcal{M}_{\theta}$ with professional cinematic judgment and the aforementioned reasoning capabilities, we fine-tune it through a two-stage paradigm using our curated expert dataset.

\subsubsection{Preference Alignment}
We first train the model on a large-scale dataset of pairwise comparisons, $\mathcal{D}_{\text{pref}} = \{ (V_{w}, V_{l}, X) \}$, where $V_w$ and $V_l$ are the preferred (win) and rejected (lose) videos, respectively. In this stage, the model learns relative cinematic aesthetics and human preferences by minimizing a Bradley-Terry ranking loss:
\begin{equation}
    \mathcal{L}_{\text{pref}}(\theta) = - \mathbb{E}_{\mathcal{D}_{\text{pref}}} \left[ \log \sigma \left( \mathcal{M}_{\theta}(V_w, X) - \mathcal{M}_{\theta}(V_l, X) \right) \right],
\end{equation}
where $\sigma$ is the sigmoid function.

\subsubsection{Score Calibration} 
To map these relative preferences into absolute, interpretable metrics and instill CoT reasoning capabilities, we subsequently fine-tune the model on a pointwise dataset $\mathcal{D}_{\text{score}} = \{ (V_i, X_i, Z_i, y_{d,i}) \}$, where $Z_i$ is the ground-truth expert CoT and $y_{d,i}$ is the absolute expert score. The model is trained to autoregressively generate the rationale $Z_i$ followed by the final score $y_{d,i}$. The optimal parameters $\theta^*$ are obtained by minimizing the Cross-Entropy loss $\mathcal{L}_{\text{CE}}$:
\begin{equation}
    \theta^* = \arg\min_{\theta} \mathbb{E}_{\mathcal{D}_{\text{score}}} \left[ \mathcal{L}_{\text{CE}} \left( \mathcal{M}_{\theta}(V, X), (Z, y_d) \right) \right].
\end{equation}

\section{Human-Machine Calibration}

\subsection{Progressive Calibration Mechanism}

To bridge the gap between demanding cinematic expert criteria and the perceptual limits of current VLMs, we propose a progressive, three-tiered calibration mechanism: (i) Prompt-Level (Rationale Replacement): Through iterative calibration, we explicitly replace evaluation dimensions and multi-questions that are overly abstract or beyond the model's perception and reasoning capabilities. (ii) Fusion-Level (Weight Optimization): To determine the exact score proportions of individual multi-questions, operator evidence ($E_{\text{prof}}$), and VLM perceptual results within the CoT, we employ a data-driven weight optimization trick. A lightweight MLP trained on human annotations to learn different weights, mitigating operator Out-of-Domain failures and VLM reasoning errors. (iii) Parameter-Level (Knowledge Injection): Fine-tuning the VLM on our expert dataset explicitly injects cinematic domain knowledge, transforming a general VLM into a specialized, expert-aligned reward model.

\begin{table*}[t]
\centering
\caption{\textbf{Human-machine alignment: pairwise win ratios.} For each video generation model and evaluation dimension, we report the pairwise win ratio against all other competitors, formatted as ``\textit{Machine Win Ratio (left) / Human Win Ratio (right)}''. The consistent correspondence between our automated predictions and expert annotations validates the efficacy of our expert-calibrated evaluation pipeline.}

\resizebox{1\linewidth}{!}{
\begin{tabular}{cccccccccccccc}
\hline

\multicolumn{2}{c}{\textbf{Evaluation Dimensions}} & \textbf{Seedance 2.0} & \textbf{Kling-v3-Omni} & \textbf{Happy Horse 1.0} & \textbf{HoloCine} & \textbf{\Centerstack{MultiShot\\Master}} & \textbf{LTX2} & \textbf{Hailuo 2.3} & \textbf{Hunyuan 1.5} & \textbf{Wan2.2} & \textbf{UniVideo} & \textbf{Vidu-Q2-Pro} \\ \hline

\multirow{2}{*}{\Centerstack{Visual Concept Design}} 
& Character & 0.61/0.63 & 0.47/0.68 & 0.74/0.82 & 0.25/0.28 & 0.25/0.05 & 0.38/0.05 & 0.48/0.89 & 0.56/0.61 & 0.42/0.39 & 0.37/0.12 & 0.61/0.56 \\ 
& Scene & 0.53/0.78 & 0.53/0.61 & 0.82/0.73 & 0.62/0.53 & 0.22/0.05 & 0.47/0.20 & 0.65/0.64 & 0.37/0.44 & 0.52/0.34 & 0.19/0.07 & 0.65/0.55 \\ 

\rowcolor[HTML]{E6F2FF} 
& Consistency & 0.79/0.81 & 0.53/0.74 & 0.58/0.66 & 0.48/0.48 & 0.22/0.20 & 0.05/0.05 & 0.90/0.70 & 0.50/0.20 & 0.47/0.40 & 0.52/0.12 & 0.39/0.29 \\ 
\rowcolor[HTML]{E6F2FF} 
& Action & 0.65/0.75 & 0.48/0.65 & 0.64/0.72 & 0.32/0.35 & 0.13/0.18 & 0.33/0.05 & 0.63/0.33 & 0.16/0.27 & 0.83/0.67 & 0.12/0.10 & 0.68/0.53 \\
\rowcolor[HTML]{E6F2FF} 
\multirow{-3}{*}{Acting} & Expression & 0.58/0.70 & 0.44/0.62 & 0.62/0.68 & 0.24/0.32 & 0.16/0.16 & 0.42/0.32 & 0.42/0.44 & 0.66/0.58 & 0.43/0.47 & 0.42/0.30 & 0.57/0.40 \\ 

\multirow{3}{*}{Cinematography} 
& Composition & 0.61/0.72 & 0.54/0.69 & 0.72/0.80 & 0.81/0.56 & 0.24/0.12 & 0.36/0.05 & 0.74/0.89 & 0.31/0.18 & 0.76/0.47 & 0.37/0.26 & 0.46/0.36 \\ 
& Pacing & 0.81/0.75 & 0.33/0.58 & 0.78/0.68 & 0.75/0.50 & 0.05/0.05 & 0.60/0.20 & 0.90/0.90 & 0.80/0.40 & 0.20/0.05 & 0.63/0.27 & 0.45/0.41 \\ 
& Lens & 0.74/0.76 & 0.48/0.67 & 0.68/0.59 & 0.58/0.44 & 0.05/0.05 & 0.46/0.16 & 0.45/0.29 & 0.44/0.39 & 0.57/0.56 & 0.40/0.12 & 0.44/0.46 \\ 

\rowcolor[HTML]{E6F2FF}  
& Visual Quality & 0.71/0.66 & 0.66/0.84 & 0.68/0.78 & 0.44/0.33 & 0.05/0.05 & 0.67/0.33 & 0.53/0.77 & 0.53/0.27 & 0.33/0.20 & 0.28/0.14 & 0.31/0.28 \\ 
\rowcolor[HTML]{E6F2FF} 
& Chromaticity & 0.73/0.76 & 0.64/0.76 & 0.72/0.60 & 0.39/0.38 & 0.13/0.05 & 0.77/0.80 & 0.23/0.05 & 0.63/0.40 & 0.37/0.80 & 0.22/0.13 & 0.38/0.35 \\ 
\rowcolor[HTML]{E6F2FF} 
& Lighting & 0.70/0.75 & 0.50/0.74 & 0.68/0.76 & 0.67/0.50 & 0.05/0.05 & 0.40/0.05 & 0.75/0.60 & 0.35/0.25 & 0.45/0.60 & 0.34/0.06 & 0.46/0.38 \\ 
\rowcolor[HTML]{E6F2FF} 
\multirow{-4}{*}{Aesthetics}
& Materiality & 0.81/0.74 & 0.65/0.77 & 0.74/0.64 & 0.51/0.33 & 0.05/0.05 & 0.55/0.10 & 0.58/0.68 & 0.47/0.43 & 0.35/0.23 & 0.23/0.11 & 0.53/0.47 \\

\multirow{2}{*}{Affectivity} 
& Grounding & 0.57/0.68 & 0.54/0.62 & 0.82/0.78 & 0.54/0.58 & 0.06/0.05 & 0.54/0.42 & 0.56/0.86 & 0.54/0.48 & 0.54/0.38 & 0.54/0.40 & 0.54/0.50 \\ 
& Progression & 0.55/0.55 & 0.53/0.70 & 0.72/0.62 & 0.53/0.65 & 0.40/0.15 & 0.48/0.28 & 0.60/0.85 & 0.42/0.22 & 0.53/0.60 & 0.48/0.32 & 0.46/0.28 \\ 	

\rowcolor[HTML]{E6F2FF} 
& Logic & 0.40/0.36 & 0.75/0.69 & 0.80/0.88 & 0.45/0.75 & 0.25/0.20 & -/- & -/- & -/- & -/- & -/- & -/- \\ 
\rowcolor[HTML]{E6F2FF} 
\multirow{-2}{*}{\Centerstack{Multi-Shot}} 
& Rhythm & 0.50/0.68 & 0.75/0.65 & 0.85/0.82 & 0.40/0.62 & 0.20/0.05 & -/- & -/- & -/- & -/- & -/- & -/- \\

\multirow{2}{*}{Sound Design} 
& Vocal & 0.45/0.58 & 0.60/0.72 & 0.85/0.72 & -/- & -/- & 0.35/0.40 & -/- & -/- & -/- & -/- & -/- \\ 
& Soundscape & 0.55/0.58 & 0.35/0.55 & 0.72/0.78 & -/- & -/- & 0.35/0.30 & -/- & -/- & -/- & -/- & -/- \\ 

\hline
\end{tabular}
}
\label{tab:win_ratio}
% \vspace{-0.1cm}
\end{table*}

\begin{table}[t]
\centering
\caption{\textbf{Human-machine alignment: correlation coefficients.} We report the Spearman Rank Correlation Coefficient (SRCC) and Pearson Linear Correlation Coefficient (PLCC), along with their respective $p$-values, between EvalVerse and human expert evaluations across all fine-grained dimensions. The consistently high correlation scores demonstrate that our automated metrics robustly align with professional human perception.}
% \vspace{-0.2cm}
\setlength{\tabcolsep}{26pt}
\resizebox{1\linewidth}{!}{
\begin{tabular}{ccccccc}
\hline

\multicolumn{2}{c}{\textbf{Evaluation Dimensions}} & \textbf{Model Number} & \textbf{SRCC} & \textbf{$p_{srcc}$} & \textbf{PLCC} & \textbf{$p_{plcc}$} \\ \hline

\multirow{2}{*}{\Centerstack{Visual Concept Design}} 
& Character & 11&+0.7529&0.0075&+0.7664&0.0059 \\ 
& Scene & 11&+0.8082&0.0026&+0.8224&0.0019 \\ 

\rowcolor[HTML]{E6F2FF} 
& Consistency & 11&+0.7472&0.0082&+0.7736&0.0052\\ 
\rowcolor[HTML]{E6F2FF} 
& Action & 11&+0.7636&0.0062&+0.7949&0.0035\\ 
\rowcolor[HTML]{E6F2FF} 
\multirow{-3}{*}{Acting} 
& Expression & 11&+0.8276&0.0017&+0.7872&0.0040\\ 

\multirow{3}{*}{Cinematography} 
& Composition & 11&+0.7545&0.0073&+0.8119&0.0024 \\ 
& Pacing & 11&+0.7517&0.0076&+0.7406&0.0091\\ 
& Lens & 11&+0.8018&0.0030&+0.7899&0.0038 \\ 

\rowcolor[HTML]{E6F2FF} 
& Visual Quality & 11&+0.7991&0.0032&+0.7875&0.0040\\ 
\rowcolor[HTML]{E6F2FF} 
& Chromaticity & 11&+0.7460&0.0084&+0.8067&0.0027\\ 
\rowcolor[HTML]{E6F2FF} 
& Lighting & 11&+0.8174&0.0021&+0.7840&0.0043 \\ 
\rowcolor[HTML]{E6F2FF} 
\multirow{-4}{*}{Aesthetics} 
& Materiality &11&+0.8091&0.0026&+0.8246&0.0018 \\ 

\multirow{2}{*}{Affectivity} 
& Grounding & 11&+0.8318&0.0015&+0.7996&0.0031\\ 
& Progression & 11&+0.8457&0.0010&+0.7634&0.0063\\ 	

\rowcolor[HTML]{E6F2FF}
& Logic & 5&+0.9000&0.0374&+0.8430&0.0729\\ 
\rowcolor[HTML]{E6F2FF}
\multirow{-2}{*}{\Centerstack{Multi-Shot}} 
& Rhythm & 5&+0.9000&0.0374&+0.8300&0.0820\\ 

\multirow{2}{*}{Sound Design} 
& Vocal & 4&+0.9487&0.0513&+0.8460&0.1540\\ 
& Soundscape & 4&+0.9487&0.0513&+0.8502&0.1498\\ \hline
\end{tabular}
}
\label{tab:correlation}
\vspace{-0.3cm}
\end{table}

\begin{figure*}[t]
    \centering
\includegraphics[width=\linewidth]{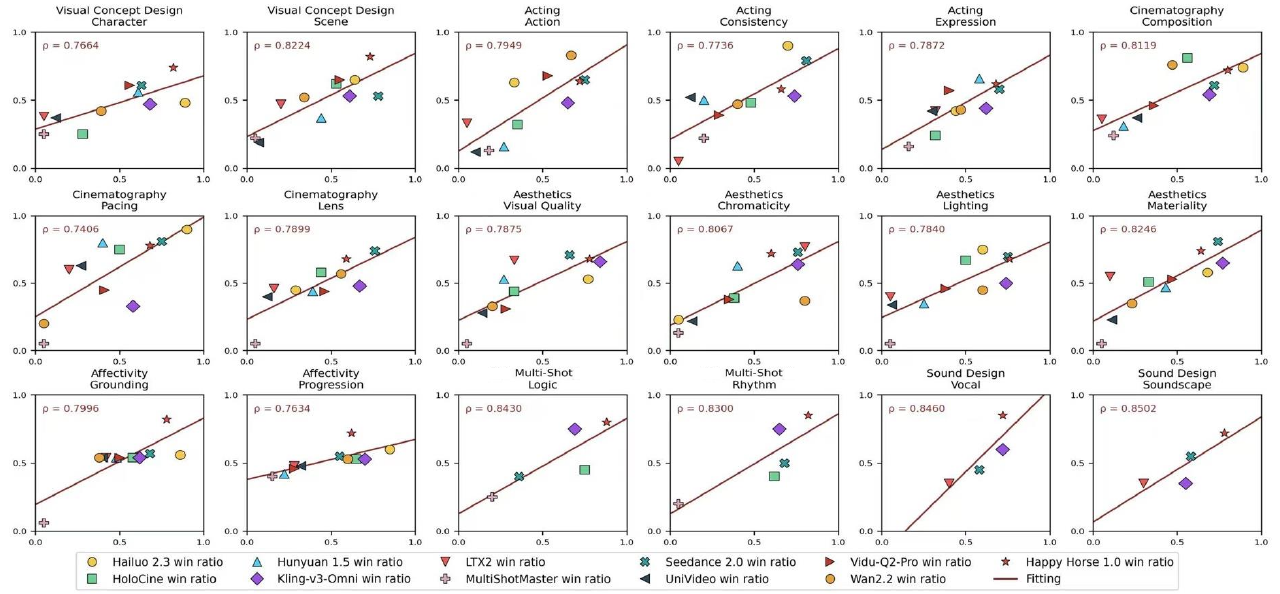}
    \vspace{-0.7cm}
    \caption{\textbf{Human-machine alignment: visualizing consistency.} Each plot correlates expert (x-axis) and machine (y-axis) win ratios per model. Linear fits and Pearson's $\rho$ confirm that EvalVerse strongly aligns with human judgment across all dimensions.}
    \label{fig:alignment}
\end{figure*}

\subsection{Alignment Analysis}

To rigorously validate the efficacy of our three-tiered calibration mechanism, we systematically assess the human-machine alignment between EvalVerse and our expert panel from three complementary perspectives: (i) granular win-ratio comparison (Tab.~\ref{tab:win_ratio}), (ii) statistical correlation analysis (Tab.~\ref{tab:correlation}), and (iii) trend consistency visualization (Fig.~\ref{fig:alignment}). Across all evaluations, we adopt the pairwise win-ratio~\cite{vbench,vbench++} as the unified comparison signal. Specifically, for every candidate model and sub-dimension, we compute its win-ratio against all competitors and measure the correlation between human-derived and EvalVerse-predicted scores. A higher correlation ($\rho$) indicates that our evaluator faithfully reproduces the relative ordering of professional human preferences.

\subsubsection{Alignment Results}

We present a comprehensive analysis of our alignment results across the three aforementioned perspectives. First, Tab.~\ref{tab:win_ratio} details the granular pairwise win-ratios, revealing a striking absolute proximity between EvalVerse predictions and expert annotations across all candidate models. Building upon this raw data, Tab.~\ref{tab:correlation} reports the per-dimension Spearman Rank Correlation Coefficient (SRCC) and Pearson Linear Correlation Coefficient (PLCC). Across all sub-dimensions, $\rho$ stays within a tight band. Finally, Fig.~\ref{fig:alignment} visualizes these relationships through scatter plots and linear regressions, where the tight linear fits further corroborate the robustness of our automated metrics. 

More importantly, the statistical results in Tab.~\ref{tab:correlation} trace out a clear pattern that is highly consistent with our design intuition: (i) Pixel-grounded dimensions (Visual Concept Design, Cinematography, Acting, Aesthetics, Affectivity), which are covered primarily by prompt-level CoT, attain strong alignment. This demonstrates that CoT-based digitization already provides a reliable backbone for the majority of cinematic criteria. (ii) Abstract and temporally-entangled dimensions, which are additionally calibrated by task-specific SFT, deliver the highest agreement with human experts. This directly verifies that parameter-level knowledge injection is the decisive step for the hardest dimensions, where natural-language rationales alone cannot close the human-machine gap.

\subsubsection{Discussion: The Complementary Synergy of CoT and SFT}

While prompt-level CoT on frozen VLMs effectively digitizes perceptually grounded dimensions (\emph{e.g.}, lighting, chromaticity) to provide broad, interpretable alignment, it fundamentally hits a perceptual ceiling for subjective, temporally-entangled, or cross-modal aspects (\emph{e.g.}, multi-shot rhythm). Abstract concepts like ``rhythmic layering'' cannot be robustly decomposed into zero-shot observable tokens, regardless of prompt elaboration. To overcome this limitation of purely verbalized prompts, we introduce task-specific SFT as a complementary calibration tier. By explicitly injecting the human scoring distribution directly into the VLM's parameters, SFT supplies the critical last-mile alignment exactly where abstract cinematic expertise lives. Rather than competing, these two paradigms synergize: CoT ensures transparent reasoning across the pipeline, while SFT bridges the perception-reasoning gap for complex dimensions.

% Together, they yield a uniformly high human-aligned evaluator, ready to serve as both a diagnostic benchmark and a reliable reward source for RL and Agentic workflows.

\section{Conclusion}

In this work, we introduced \textbf{EvalVerse}, fundamentally redefining video generation assessment from basic prompt-following (``whether it is right'') to a rigorous audit of professional filmmaking (``whether it is good''). By structurally mirroring the real-world pipeline and proposing a systematic human-machine calibration mechanism, we provide a principled framework for characterizing and injecting nuanced human preferences into algorithmic scoring. This successfully digitizes subjective expertise into computable metrics, bridging the long-standing credibility gap between human aesthetic perception and machine evaluation. Extending far beyond a static leaderboard, EvalVerse establishes a foundational infrastructure for the post-SFT era by supplying dense, expert-aligned reward vectors for Reinforcement Learning and explainable diagnostic feedback for autonomous agentic workflows. By providing this critical ``missing link,'' it catalyzes the transformation of generative models from passive clip generators into professional-grade virtual directors, ushering in a new era of computable cinematography for computer graphics community.

% \noindent
% \textcolor{red}{\textbf{More Details in the Supplementary Material.} All necessary details essential for full reproducibility are provided in the supplementary document. This includes the definitions and criteria for our cinematic taxonomy, the deployment setups for the automated evaluation pipeline, as well as dataset statistics and implementation specifics.}

\noindent
\textbf{Limitations and Future Work.} Future work will address several key challenges: (i) \textit{VLM Bottlenecks:} Current VLMs process discrete keyframes rather than continuous streams, limiting its temporal perception; (ii) \textit{Long-Form Narratives:} Scaling evaluation to macro-narratives (\emph{e.g.}, 10+ minutes) requires advanced long-context reasoning; and (iii) \textit{Artistic Diversity:} Assessing boundless avant-garde styles remains difficult. Ultimately, natively integrating ``evaluation'' as a fundamental ``understanding'' task into unified multi-modal models represents a highly promising frontier.

% \noindent\textbf{Future Work: From Static Benchmark to Dynamic Infrastructure.} 
% While this paper establishes EvalVerse as a comprehensive diagnostic benchmark, its pipeline-aware architecture and expert-calibrated scoring unlock significant potential for active generative workflows. In future work, we aim to transition EvalVerse from a static evaluator to a dynamic infrastructure. First, we plan to integrate EvalVerse as an off-the-shelf, domain-specific Reward Model (RM) for Reinforcement Learning paradigms (e.g., RLHF~\cite{rlhf} and GRPO~\cite{dancegrpo}). Unlike conventional monolithic evaluators that often induce ``reward hacking'' (e.g., generating static, highly detailed images to maximize visual scores), EvalVerse provides dense, expert-aligned reward vectors. This will allow RL algorithms to optimize for specific cinematic weaknesses—such as penalizing physical violations or rewarding emotional-action synergy—without compromising dynamic motion. Furthermore, we will explore its application as an ``Environment Critic'' within autonomous Agentic workflows. By leveraging its stage-by-stage diagnostic rationales, LLM-driven virtual directors can parse explainable, natural language feedback to iteratively adjust tool calls and prompt rewrites, effectively closing the self-correction loop for long-form cinematic creation.

% ---- Bibliography ----

\bibliographystyle{plainnat}
\bibliography{sample-base}

% (Left) The rigorous annotation process, where raw videos are processed by industrial operators and verified via human checking to generate highly structured JSON metadata. (Top Right) The pie charts illustrate the proportional distribution used to ensure balanced and comprehensive data sampling. (Bottom) The test pair construction module which generates multi-modal inputs to support a wide range of downstream generation tasks.

\end{document}